\documentclass[10pt,twocolumn,letterpaper]{article}

\usepackage[pagenumbers]{wacv} 

\usepackage{newtxtext}
\definecolor{wacvblue}{rgb}{0.21,0.49,0.74}
\usepackage[pagebackref,breaklinks,colorlinks,allcolors=wacvblue]{hyperref}
\usepackage[most]{tcolorbox}
\usepackage{makecell}
\usepackage{hyperref}
\usepackage{url}
\usepackage{bbm}
\usepackage{caption}
\usepackage{quickmath}
\usepackage{lipsum}
\usepackage[ruled,vlined]{algorithm2e}
\SetKwComment{Comment}{/*}{*/}
\usepackage{array}
\usepackage{soul}
\usepackage{titletoc}
\usepackage{fontawesome5}

\usepackage[table]{xcolor}
\definecolor{lightpurple}{RGB}{180,200,230}
\definecolor{linkblue}{RGB}{70,130,180}
\sethlcolor{lightpurple!25}
\usepackage{enumitem,amssymb}
\usepackage{wrapfig}
\usepackage{multirow}  
\usepackage{graphicx}
\newlist{todolist}{itemize}{2}
\setlist[todolist]{label=$\square$}
\usepackage{pifont}
\usepackage{multirow}
\usepackage{tabularx,threeparttable,array}
\newcommand{\cmark}{\ding{51}}%
\newcommand{\xmark}{\ding{55}}%

\usepackage{algorithmic}
\usepackage{xspace}
\usepackage{makecell}
\usepackage{pifont}

\newcolumntype{P}{>{\centering\arraybackslash}X}

\newcolumntype{Y}{>{\centering\arraybackslash}X}

\newcolumntype{Z}{>{\columncolor{lightpurple!25}\centering\arraybackslash}X}
\newcommand{\std}[1]{\,\scriptsize{\textcolor{gray}{±#1}}}

\newcommand{\ours}{\textsc{AlignCP}\xspace}

\renewcommand{\arraystretch}{1}

\definecolor{Gray}{gray}{0.9}
\definecolor{Better}{rgb}{0.18, 0.407, 0.266}
\definecolor{Worse}{rgb}{0.35, 0.35, 0.35}
\definecolor{drakgreen}{rgb}{0.38, 0.67, 0.38}
\definecolor{drakpurple}{rgb}{0.38, 0.27, 0.61}
\definecolor{granate}{rgb}{0.64, 0.16, 0.16}

\newcommand{\imp}[1]{$_{{\textbf{\textcolor{Better}{#1}}}}$}
\newcommand{\wor}[1]{$_{{\textbf{\textcolor{Worse}{#1}}}}$}

\def\wacvPaperID{1649} 
\def\confName{WACV}
\def\confYear{2027}

\title{Learning to Adapt and Calibrate: Score Distribution Alignment for Few-Shot Uncertainty Prediction in Medical VLMs}

\author{
Xuan Cuong Ngo \qquad Ngan Le\\
University of Arkansas\\
Fayetteville, Arkansas, USA\\
{\tt\small \{cngo,thile\}@uark.edu}
}

\begin{document}
\maketitle
\begin{abstract}

Uncertainty estimation for medical vision--language models (VLMs) using conformal prediction has gained increasing attention due to its distribution-free coverage guarantees under exchangeability, but applying it in few-shot transfer is challenging. In this setting, only a small labeled support set is available, yet both supervised adaptation and conformal calibration require labeled data. A natural strategy is therefore to use the support set for supervised fine-tuning and then reuse it for conformal calibration. However, once the model has been adapted on these samples, their nonconformity scores are no longer exchangeable with those of unseen query samples, and naively applying split conformal prediction (SCP) can lead to unreliable coverage. Existing transductive conformal adaptation methods often avoid supervised updates to preserve the standard conformal protocol, but this sacrifices much of the value of few-shot supervision. In few-shot transfer, adaptation is the primary objective; conformal prediction should instead serve as an uncertainty estimation layer for the adapted model rather than restricting adaptation itself.
We study this problem from the perspective of adaptation-induced non-exchangeability and propose \textbf{\ours}, a framework that reconciles supervised few-shot adaptation with conformal uncertainty estimation. \ours first adapts a medical VLM on the labeled support set and then reuses the same samples for calibration. To compensate for the resulting mismatch between support and query scores, \ours uses unlabeled query predictions to reweight the support nonconformity scores, producing a corrected calibration distribution from which the conformal threshold is computed. Since supervised adaptation breaks the exchangeability required by standard SCP, \ours does not claim its standard marginal coverage guarantee; instead, it aims to recover reliable empirical coverage as closely as possible by correcting the adaptation-induced score-distribution shift without requiring query labels.
Extensive experiments across diverse medical domains show that \ours improves predictive performance over unsupervised conformal baselines while maintaining reliable empirical coverage and competitive prediction-set efficiency under distribution shift.

\end{abstract}

\section{Introduction}
\label{intro}


Medical vision–language models (VLMs) have recently demonstrated strong transfer capabilities across a range of clinical applications, including image classification \cite{du2025medical}, visual question answering \cite{liu2023q2atransformer}, and cross-modal reasoning. As a result, these models are increasingly adopted in high-stakes healthcare settings, with specialized medical VLMs developed for radiology \cite{zhang2022contrastive,wang2022medclip}, histology \cite{huang2023visual,lu2024visual}, and retinal imaging \cite{silva2025foundation}. Their effectiveness largely stems from large-scale pretraining, which enables strong generalization even when labeled medical data is limited. However, despite these promising advances, the reliability and uncertainty behavior of medical VLMs under transfer remain insufficiently studied, with only a few recent works beginning to address this issue \cite{silva2025trustworthy}.

Split conformal prediction (SCP) provides a principled, distribution-free framework for uncertainty quantification, offering finite-sample coverage guarantees under the exchangeability assumption \cite{gammerman2013learning,vovk2005algorithmic}. Recent works have extended SCP to neural networks, showing that reliable prediction sets can be obtained using a small labeled calibration set \cite{angelopoulos2020uncertainty,ding2023class}. However, in transfer learning scenarios, it is common practice to perform supervised fine-tuning on available support data to improve predictive performance \cite{radford2021learning,huang2024lp++}. When SCP is applied after such adaptation, the predictive model has already been updated using part of the target data. This violates the exchangeability assumption between calibration and test samples, thereby undermining the theoretical coverage guarantees of SCP \cite{silva2025trustworthy}.

To mitigate this issue, recent transductive conformal adaptation methods propose avoiding supervised updates and instead performing unsupervised adaptation jointly over calibration and test covariates \cite{silva2025trustworthy}. These methods are designed to preserve empirical coverage by preventing label-driven shifts in the nonconformity score distribution. Nevertheless, this strategy is misaligned with the central goal of few-shot learning. In few-shot transfer, the main objective is to exploit the limited labeled support set as effectively as possible to adapt the model to the target task. Uncertainty estimation through conformal prediction should assist this adapted model by providing a calibrated view of its predictions, rather than becoming the constraint that prevents supervised adaptation. Therefore, while transductive conformal adaptation can maintain validity, it may do so by prioritizing conformal assumptions over task adaptation, effectively leaving valuable labeled supervision underused. This trade-off is especially undesirable in medical few-shot settings, where a small number of labeled examples can provide critical task-specific information and significantly improve predictive accuracy \cite{hou2019cross}.

In this work, we develop a principled framework for supervised few-shot adaptation of medical VLMs with reliable conformal uncertainty estimation under distribution shift. Unlike transductive conformal adaptation methods that preserve exchangeability by avoiding supervised updates, we adopt a perspective that is more consistent to the goal of few-shot learning. In few-shot transfer, the scarce labeled support samples provide the most direct supervision for adapting a pretrained model to the target medical task; therefore, task adaptation should be the first priority. Conformal prediction should then serve as an auxiliary uncertainty estimation mechanism for the adapted model, rather than a constraint that prevents supervised learning from the available labels. However, once the model is fine-tuned on the support set, the calibration and test nonconformity scores are no longer exchangeable, causing standard conformal prediction to suffer from undercoverage. For this reason, our objective is not to guarantee the exact marginal coverage of standard conformal prediction, but to close the resulting coverage gap and provide the most trustworthy uncertainty estimates possible for the adapted medical VLM.

We begin by revisiting supervised few-shot conformal transfer from a distributional perspective. In this setting, the limited labeled support set is used both to fine-tune the model for the target task and to provide the calibration scores for conformal prediction. While supervised fine-tuning improves task adaptation, it can also bias the adapted model toward the support/calibration samples, shifting the induced nonconformity score distribution away from that of the test data. Since the coverage behavior of split conformal prediction depends on the calibration scores being representative of the test scores, this score-level mismatch can directly lead to coverage degradation. Recent theoretical results formalize this phenomenon by bounding the coverage gap in terms of discrepancies between the pushforward score distributions under calibration and test data \cite{wang2025non}. These insights suggest that the central challenge is to control the score distribution shift induced by supervised adaptation, so that conformal prediction can provide reliable uncertainty estimates for the adapted model under distribution shift.

Motivated by this insight, we propose \ours, a framework that enables supervised few-shot fine-tuning while explicitly reducing the score distribution shift induced by adaptation. After adapting the VLM on the labeled support set, \ours learns a reweighted calibration score distribution that better matches the unlabeled test score distribution. This is achieved by minimizing a theoretically motivated objective derived from a coverage-gap bound defined over the two score distributions. Since supervised adaptation violates the exchangeability assumption through score-level distribution mismatch, reducing this mismatch helps narrow the resulting coverage gap and improve the reliability of conformal uncertainty estimates. Importantly, because nonconformity scores are one-dimensional, the proposed alignment is computationally efficient and remains stable in data-scarce few-shot regimes.

Extensive experiments across diverse medical domains demonstrate that \ours achieves stronger predictive performance than purely unsupervised transductive approaches, while maintaining reliable empirical coverage and competitive prediction set efficiency under distribution shift.

The main contributions are summarized as follows:
\begin{itemize}
    \item[$\circ$] \textbf{Supervised Few-Shot Conformal Adaptation:} We propose \ours, a principled framework for uncertainty estimation in supervised few-shot adaptation of medical VLMs. Unlike transductive approaches that avoid supervised updates, \ours first adapts the model using the limited labeled support set and then corrects the induced score-level mismatch.
    
    \item[$\circ$] \textbf{Coverage Gap Minimization Perspective:} We reformulate conformal prediction after supervised few-shot adaptation as a coverage gap minimization problem. Building on recent theoretical bounds, \ours learns a reweighted calibration score distribution that better matches the unlabeled test score distribution by minimizing a theoretically motivated objective defined over the two score distributions.
    
    \item[$\circ$] \textbf{Improved Accuracy-Coverage Trade-off:} Extensive experiments across diverse medical domains demonstrate that \ours achieves stronger predictive performance than purely unsupervised transductive methods, while maintaining reliable empirical coverage and competitive prediction set efficiency.

\end{itemize}

\section{Related Work}
\label{sec:related_work}

\noindent
\textbf{Conformal prediction for classification.}
Conformal prediction constructs prediction sets with finite-sample marginal coverage under exchangeability~\cite{vovk2005algorithmic,shafer2008tutorial,angelopoulos2024conformal}. For classification, common nonconformity scores include least ambiguous classification (LAC)~\cite{sadinle2019least}, adaptive prediction sets (APS)~\cite{romano2020classification}, and regularized adaptive prediction sets (RAPS)~\cite{angelopoulos2020uncertainty}. These methods improve the reliability of classifier outputs by calibrating prediction sets on held-out data. However, their coverage guarantee relies on calibration and test samples following the same score distribution, which is often violated under domain shift.

\noindent
\textbf{Conformal prediction under distribution shift.}
Several works study conformal prediction beyond the standard i.i.d. setting. Weighted conformal prediction reweights calibration samples to address covariate shift~\cite{tibshirani2019conformal}, while more general approaches analyze conformal validity under distribution drift and non-exchangeability~\cite{barber2023conformal}. In medical imaging, distribution shift is especially common because datasets can differ in imaging protocol, patient population, disease prevalence, and annotation process. Recent studies show that conformal prediction can suffer from degraded coverage under such shifts~\cite{mehrtens2023pitfalls}. Unlike prior reweighting-based methods, \ours{} does not assume any prior knowledge about the source-target distribution shift.

\noindent
\textbf{Few-shot adaptation and transductive learning.}
Few-shot adaptation improves target-domain performance using a small labeled support set, while transductive methods additionally use unlabeled test samples during inference. TIM optimizes mutual information over query predictions~\cite{boudiaf2020information}, and TransCLIP extends transductive inference to VLMs~\cite{zanella2024boosting}. SCA-T applies transductive optimization to conformal prediction for few-shot medical VLMs~\cite{silva2025trustworthy}.These methods either avoid supervised adaptation to preserve calibration, or apply adaptation in a way that can change the calibration score distribution and cause undercoverage. In contrast, \ours{} explicitly combines supervised adaptation with score-distribution alignment, retaining the accuracy gain from adaptation while correcting the calibration mismatch needed for reliable conformal prediction.

\noindent
\textbf{Medical VLMs.}
VLMs such as CLIP learn transferable image-text representations through contrastive learning~\cite{radford2021learning}. This paradigm has been adapted to medical imaging through methods such as ConVIRT~\cite{zhang2022contrastive}, GLoRIA~\cite{huang2021gloria}, BioViL~\cite{boecking2022making}, MedCLIP~\cite{wang2022medclip}, and CheXzero~\cite{tiu2022expert}. These models provide strong zero-shot and few-shot transfer ability, but their predictions can still be miscalibrated when deployed on shifted target datasets. \ours{} is complementary to these representation-learning methods, focusing on reliable adaptation and conformal calibration for medical VLMs.

\section{Methodology}
\subsection{Notation and Problem Setup}
\label{sec:notation}

We first introduce the notation used throughout the method section. 
Let $\mathcal{X}$ denote the input space and $\mathcal{Y}$ denote the label space. 
We use uppercase letters, e.g., $(X,Y)$, for random variables and lowercase letters, e.g., $(x,y)$, for their realizations. 
For a distribution $R$ over $\mathcal{X}\times\mathcal{Y}$, $R_X$ denotes its marginal distribution over $\mathcal{X}$, and $R^n$ denotes the $n$-fold product distribution obtained by drawing $n$ independent samples from $R$.

We consider a source distribution $P$ and a target distribution $Q$ over $\mathcal{X}\times\mathcal{Y}$. 
The labeled calibration set is drawn from the source distribution,
\begin{equation}
    \mathcal{D}_{\mathrm{cal}}
    =
    \{(X_i,Y_i)\}_{i=1}^{n}
    \sim P^n,
\end{equation}
while the unlabeled target query set is drawn from the target marginal distribution,
\begin{equation}
    \mathcal{D}_{\mathrm{qry}}
    =
    \{\widetilde X_j\}_{j=1}^{m}
    \sim Q_X^m.
\end{equation}

Let $f_{\theta}$ denote a pretrained VLM, and let $\theta'$ denote the parameters after supervised adaptation. 
The nonconformity score is written as $s_{\theta'}:\mathcal{X}\times\mathcal{Y}\rightarrow\mathbb{R}$, where larger scores indicate lower conformity between an input and a candidate label. 
For simplicity, we write $s(x,y)$ instead of $s_{\theta'}(x,y)$ when the model parameters are clear from context.

For any distribution $R$ over $\mathcal{X}\times\mathcal{Y}$, the induced score random variable is $S_R=s(X,Y)$ for $(X,Y)\sim R$. 
The corresponding score distribution is the pushforward distribution $R_s:=s_{\#}R$. 
Thus, source examples induce $P_s=s_{\#}P$, while target examples induce $Q_s=s_{\#}Q$. 
We denote the cumulative distribution function of $R_s$ by $F_{R_s}$ and, when it exists, its density by $p_{R_s}$.

For each calibration sample, the calibration score is $S_i=s(X_i,Y_i)$, and the set of calibration scores is denoted by $\mathcal{S}_{\mathrm{cal}}=\{S_i\}_{i=1}^{n}$. 
Since $\mathcal{D}_{\mathrm{cal}}\sim P^n$, the induced calibration scores satisfy $\mathcal{S}_{\mathrm{cal}}\sim P_s^n$. 
Here, $P^n$ denotes the distribution of source data pairs, whereas $P_s^n$ denotes the distribution of their induced nonconformity scores.

Given $\mathcal{S}_{\mathrm{cal}}$, we denote the split conformal threshold at miscoverage level $\alpha\in(0,1)$ by $\widehat q_{1-\alpha}(\mathcal{S}_{\mathrm{cal}})$. 
The corresponding prediction set is
\begin{equation}
    \mathcal{C}_{\alpha}(x)
    =
    \left\{
    y\in\mathcal{Y}:
    s(x,y)
    \leq
    \widehat q_{1-\alpha}(\mathcal{S}_{\mathrm{cal}})
    \right\}.
\end{equation}
For a test example $(X_t,Y_t)$, we write its nonconformity score as $S_t=s(X_t,Y_t)$.

Finally, $\operatorname{Cov}_{P}(\alpha)$ and $\operatorname{Cov}_{Q}(\alpha)$ denote the marginal coverage of the same conformal procedure when the test point is drawn from $P$ and $Q$, respectively. 
The coverage gap between the source and target distributions is
\begin{equation}
    \Delta_{P,Q}(\alpha)
    =
    \left|
    \operatorname{Cov}_{P}(\alpha)
    -
    \operatorname{Cov}_{Q}(\alpha)
    \right|.
\end{equation}
This quantity will be used to characterize how distribution shift in score space affects conformal coverage.

\subsection{Revisiting Conformal Prediction}
\label{sec:revisiting-cp}

Conformal prediction (CP)~\cite{vovk2005algorithmic} constructs prediction sets with finite-sample coverage guarantees under exchangeability. 
We first consider the standard setting where the calibration and test examples are drawn from the same distribution $P$ over $\mathcal{X}\times\mathcal{Y}$. 
Given calibration scores $\mathcal{S}_{\mathrm{cal}}=\{S_i\}_{i=1}^{n}\sim P_s^n$, split conformal prediction (SCP)~\cite{papadopoulos2002inductive} chooses a threshold from the empirical quantile of the calibration scores.

Let $S_{(1)}\leq \cdots \leq S_{(n)}$ denote the order statistics of $\mathcal{S}_{\mathrm{cal}}$. 
For a target miscoverage level $\alpha\in(0,1)$, the split conformal threshold is
\begin{equation}
    \widehat q_{1-\alpha}(\mathcal{S}_{\mathrm{cal}})
    :=
    S_{\left(\left\lceil (n+1)(1-\alpha) \right\rceil\right)},
\end{equation}
with the convention that $S_{(n+1)}=\infty$ when the index exceeds $n$. 
The prediction set for a test input $x$ is then
\begin{equation}
    \mathcal{C}_{\alpha}(x)
    =
    \left\{
    y\in\mathcal{Y}:
    s(x,y)
    \leq
    \widehat q_{1-\alpha}(\mathcal{S}_{\mathrm{cal}})
    \right\}.
\end{equation}

For a test example $(X_t,Y_t)\sim P$, let $S_t=s(X_t,Y_t)$ denote its nonconformity score. 
Since $S_t$ and the calibration scores are drawn from the same score distribution $P_s$, SCP guarantees
\begin{equation}
\begin{aligned}
    \mathbb{P}
    \bigl(
    Y_t \in \mathcal{C}_{\alpha}(X_t)
    \bigr)
    &=
    \mathbb{P}
    \left(
    S_t
    \leq
    \widehat q_{1-\alpha}(\mathcal{S}_{\mathrm{cal}})
    \right)  \\
    &\geq
    1-\alpha .
\end{aligned}
\end{equation}
This formulation shows that conformal validity depends on the relationship between the calibration score distribution and the test score distribution. 
This perspective becomes central under distribution shift, where calibration scores follow $P_s$ but target test scores follow $Q_s$.

\subsection{Coverage Gap under Distribution Shift}
\label{sec:coverage-gap}

We now consider the distribution-shift setting, where the calibration scores are drawn from the source score distribution $P_s$, while target test scores are drawn from $Q_s$. 
The conformal threshold $\widehat q_{1-\alpha}(\mathcal{S}_{\mathrm{cal}})$ is still computed from the source calibration scores $\mathcal{S}_{\mathrm{cal}}\sim P_s^n$. 
However, when evaluating on the target distribution, the test score follows $S_t\sim Q_s$. 
Therefore, if $P_s\neq Q_s$, the source calibration scores and target test scores are no longer drawn from the same score distribution, and the standard split conformal guarantee under $P$ does not directly transfer to $Q$.

Using the same source-calibrated threshold, the source coverage can be written as
\begin{equation}
\begin{aligned}
    \operatorname{Cov}_{P}(\alpha)
    =
    \mathbb{E}_{\mathcal{S}_{\mathrm{cal}}\sim P_s^n}
    \left[
        \mathbb{E}_{S_t\sim P_s}
        \left[
            \mathbf{1}
            \left\{
            S_t
            \leq
            \widehat q_{1-\alpha}(\mathcal{S}_{\mathrm{cal}})
            \right\}
        \right]
    \right].
\end{aligned}
\end{equation}
Similarly, the target coverage is
\begin{equation}
\begin{aligned}
    \operatorname{Cov}_{Q}(\alpha)
    =
    \mathbb{E}_{\mathcal{S}_{\mathrm{cal}}\sim P_s^n}
    \left[
        \mathbb{E}_{S_t\sim Q_s}
        \left[
            \mathbf{1}
            \left\{
            S_t
            \leq
            \widehat q_{1-\alpha}(\mathcal{S}_{\mathrm{cal}})
            \right\}
        \right]
    \right].
\end{aligned}
\end{equation}
Thus, both quantities use the same random threshold learned from $P_s$, but differ in the distribution used to evaluate the test score. 
The coverage gap $\Delta_{P,Q}(\alpha)$ therefore captures the change in coverage caused by replacing the source test score distribution $P_s$ with the target test score distribution $Q_s$.

Under the source distribution, split conformal prediction guarantees
\begin{equation}
    \operatorname{Cov}_{P}(\alpha)
    \geq
    1-\alpha.
\end{equation}
For the target distribution, the definition of the coverage gap gives
\begin{equation}
    \operatorname{Cov}_{Q}(\alpha)
    \geq
    \operatorname{Cov}_{P}(\alpha)
    -
    \Delta_{P,Q}(\alpha).
\end{equation}
Combining the two inequalities yields
\begin{equation}
    \operatorname{Cov}_{Q}(\alpha)
    \geq
    1-\alpha-\Delta_{P,Q}(\alpha).
    \label{eq:cov_guarantee}
\end{equation}
This inequality shows that target coverage degrades according to the coverage gap induced by the mismatch between $P_s$ and $Q_s$. 
Consequently, improving target reliability under distribution shift can be viewed as reducing the score-level discrepancy that contributes to $\Delta_{P,Q}(\alpha)$.

\subsection{Supervised Conformal Transfer under Distribution Shift}
\label{sec:supervised-conformal-transfer}

Building on the coverage-gap view above, our goal is to improve target coverage after supervised adaptation by reducing the score-level mismatch between the source calibration distribution and the target test distribution. 
Since the target labels are unavailable, we cannot directly observe target scores from $Q_s$. 
Instead, \ours learns a reweighted calibration score distribution from the labeled source calibration set and the unlabeled target query inputs.

\noindent
\textbf{Reweighted calibration distribution.}
To adjust the source calibration scores toward the target score distribution, \ours assigns a nonnegative weight $w_i$ to each calibration score $S_i$, with $w_i\geq 0$ and $\sum_{i=1}^{n}w_i=1$. 
These weights define the weighted empirical calibration score distribution
\begin{equation}
    P_{s,n}^{w}
    =
    \sum_{i=1}^{n}
    w_i\delta_{S_i},
\end{equation}
where $\delta_{S_i}$ denotes the Dirac measure at $S_i$. 
When $w_i=1/n$ for all $i$, $P_{s,n}^{w}$ reduces to the standard unweighted empirical calibration score distribution. 
The corresponding weighted empirical CDF is
\begin{equation}
    F_{P_{s,n}^{w}}(t)
    =
    \sum_{i=1}^{n}
    w_i\mathbf{1}\{S_i\leq t\}.
\end{equation}
The weighted conformal threshold is defined as
\begin{equation}
    \widehat q^{\,w}_{1-\alpha}
    :=
    \inf
    \left\{
    t\in\mathbb{R}:
    F_{P_{s,n}^{w}}(t)
    \geq
    1-\alpha
    \right\}.
\end{equation}
This gives the weighted prediction set
\begin{equation}
    \mathcal{C}^{w}_{\alpha}(x)
    =
    \left\{
    y\in\mathcal{Y}:
    s(x,y)
    \leq
    \widehat q^{\,w}_{1-\alpha}
    \right\}.
\end{equation}
Thus, the central problem is to choose weights such that $P_{s,n}^{w}$ better matches the target score distribution and reduces the induced coverage gap.

\noindent
\textbf{Coverage gap as score distribution discrepancy.}
The previous subsection shows that target coverage is controlled by $\Delta_{P,Q}$. 
Under mild regularity conditions, such as the existence of a density for $P_s$, the total coverage gap can be bounded by a discrepancy between the source and target score CDFs \cite{correia2026non}:
\begin{equation}
    \Delta_{P,Q}
    \leq
    \int_{\mathbb{R}}
    p_{P_s}(u)
    \left|
    F_{P_s}(u)
    -
    F_{Q_s}(u)
    \right|
    \,du .
    \label{eq:density-bound}
\end{equation}
A proof is provided in the appendix. 
This bound motivates aligning calibration and target distributions directly in the one-dimensional score space.

\noindent
\textbf{Approximating the target score distribution from unlabeled queries.}
The bound in~\eqref{eq:density-bound} depends on $Q_s$, which cannot be directly evaluated without target labels. 
However, in classification, we can compute the nonconformity score for every candidate label even when the true label is unknown. 
For each unlabeled query input $\widetilde X_j$, we define the lower and upper candidate scores as
\begin{equation}
    S_j^{\downarrow}
    =
    \min_{y\in\mathcal{Y}}s(\widetilde X_j,y),
    \qquad
    S_j^{\uparrow}
    =
    \max_{y\in\mathcal{Y}}s(\widetilde X_j,y).
\end{equation}
These scores induce two empirical surrogate distributions,
\begin{equation}
    \widehat Q_{s,m}^{\downarrow}
    =
    \frac{1}{m}
    \sum_{j=1}^{m}
    \delta_{S_j^{\downarrow}},
    \qquad
    \widehat Q_{s,m}^{\uparrow}
    =
    \frac{1}{m}
    \sum_{j=1}^{m}
    \delta_{S_j^{\uparrow}}.
\end{equation}
For brevity, we write $\widehat Q^{\downarrow}:=\widehat Q_{s,m}^{\downarrow}$ and $\widehat Q^{\uparrow}:=\widehat Q_{s,m}^{\uparrow}$.

Let $\widetilde Y_j\sim Q(\cdot\mid \widetilde X_j)$ denote the unknown target label and let $\widetilde S_j=s(\widetilde X_j,\widetilde Y_j)$ be the corresponding unobserved target score. 
By construction, $S_j^{\downarrow}\leq \widetilde S_j\leq S_j^{\uparrow}$ for every query sample. 
If the target labels were known, the empirical target score distribution would be $\widehat Q_s=\frac{1}{m}\sum_{j=1}^{m}\delta_{\widetilde S_j}$. 
Although $\widehat Q_s$ is unobserved, the lower and upper surrogate scores bracket each unobserved target score. 
Consequently, their empirical CDFs satisfy
\begin{equation}
    F_{\widehat Q^{\uparrow}}(t)
    \leq
    F_{\widehat Q_s}(t)
    \leq
    F_{\widehat Q^{\downarrow}}(t),
    \qquad
    \forall t\in\mathbb{R}.
    \label{eq:cdf-bracketing}
\end{equation}
This bracketing relation provides a computable way to approximate the unknown target score distribution using only unlabeled query inputs.

\noindent
\textbf{Sandwich upper bound.}
Directly minimizing the discrepancy in~\eqref{eq:density-bound} is infeasible because $F_{Q_s}$ is unknown. 
We therefore derive a computable surrogate using the lower and upper query-induced CDFs. 
For any surrogate distribution $\widehat Q$ and any $u\in\mathbb{R}$, the triangle inequality gives

\begin{equation}
\begin{aligned}
    \left|
    F_{P_s}(u)
    -
    F_{\widehat Q_s}(u)
    \right|
    &\leq
    \left|
    F_{P_s}(u)
    -
    F_{\widehat Q}(u)
    \right|
    \\
    &\quad+
    \left|
    F_{\widehat Q_s}(u)
    -
    F_{\widehat Q}(u)
    \right|.
\end{aligned}
\label{eq:triangle}
\end{equation}

Applying~\eqref{eq:triangle} with $\widehat Q^{\uparrow}$ and $\widehat Q^{\downarrow}$, and then averaging the two inequalities, gives
\begin{equation}
\begin{aligned}
    &
    \left|F_{P_s}(u)-F_{\widehat Q_s}(u)\right| \\
    &\leq
    \frac{1}{2}
    \Big(
    \left|F_{P_s}(u)-F_{\widehat Q^{\uparrow}}(u)\right|
    +
    \left|F_{P_s}(u)-F_{\widehat Q^{\downarrow}}(u)\right|
    \\
    &\quad+
    \left|F_{\widehat Q_s}(u)-F_{\widehat Q^{\uparrow}}(u)\right|
    +
    \left|F_{\widehat Q_s}(u)-F_{\widehat Q^{\downarrow}}(u)\right|
    \Big).
\end{aligned}
\label{eq:avg-bound}
\end{equation}

Using the bracketing relation in~\eqref{eq:cdf-bracketing}, the last two terms reduce to the bracket width:
\begin{equation}
\begin{aligned}
    &
    \left|
    F_{\widehat Q_s}(u)
    -
    F_{\widehat Q^{\uparrow}}(u)
    \right|
    +
    \left|
    F_{\widehat Q_s}(u)
    -
    F_{\widehat Q^{\downarrow}}(u)
    \right|
    \\
    &=
    F_{\widehat Q^{\downarrow}}(u)
    -
    F_{\widehat Q^{\uparrow}}(u).
\end{aligned}
\end{equation}

This yields the sandwich-type upper bound
\begin{equation}
\begin{aligned}
    &
    \int_{\mathbb{R}}
    p_{P_s}(u)
    \left|
    F_{P_s}(u)
    -
    F_{\widehat Q_s}(u)
    \right|
    \,du
    \\
    &\leq
    \frac{1}{2}
    \int_{\mathbb{R}}
    p_{P_s}(u)
    \Big(
    \left|
    F_{P_s}(u)
    -
    F_{\widehat Q^{\uparrow}}(u)
    \right|
    +
    \left|
    F_{P_s}(u)
    -
    F_{\widehat Q^{\downarrow}}(u)
    \right|
    \\
    &\hspace{5.2em}
    +
    F_{\widehat Q^{\downarrow}}(u)
    -
    F_{\widehat Q^{\uparrow}}(u)
    \Big)
    \,du .
\end{aligned}
\label{eq:sandwich-bound}
\end{equation}
The first two terms encourage the source score CDF to align with the two query-induced surrogate CDFs, while the last term accounts for the uncertainty interval between the lower and upper surrogate distributions.

\noindent
\textbf{Empirical objective.}
In practice, \ours replaces the population source score CDF $F_{P_s}$ with the weighted empirical calibration CDF $F_{P_{s,n}^{w}}$. 
We evaluate the discrepancy on the fixed calibration score support, which gives
\begin{equation}
\begin{aligned}
    \mathcal{L}_{\mathrm{align}}(w)
    &=
    \frac{1}{2n}
    \sum_{i=1}^{n}
    \Big(
    \left|
    F_{P_{s,n}^{w}}(S_i)
    -
    F_{\widehat Q^{\uparrow}}(S_i)
    \right|
    \\
    &\hspace{-2.5em}
    +
    \left|
    F_{P_{s,n}^{w}}(S_i)
    -
    F_{\widehat Q^{\downarrow}}(S_i)
    \right|
    +
    F_{\widehat Q^{\downarrow}}(S_i)
    -
    F_{\widehat Q^{\uparrow}}(S_i)
    \Big),
\end{aligned}
\label{eq:empirical-align-objective}
\end{equation}

subject to $w_i\geq 0$ and $\sum_{i=1}^{n}w_i=1$. 
Here, the weights affect the objective through the weighted CDF $F_{P_{s,n}^{w}}$, rather than through the outer summation. 
Minimizing $\mathcal{L}_{\mathrm{align}}(w)$ therefore learns a reweighted calibration distribution whose score CDF is closer to the query-induced surrogate target distributions. 
The learned distribution $P_{s,n}^{w}$ is then used to compute $\widehat q^{\,w}_{1-\alpha}$ and construct the final weighted prediction set $\mathcal{C}^{w}_{\alpha}(x)$.

\noindent
\textbf{Why minimizing Eq.~\ref{eq:empirical-align-objective} helps ?}
Eq.~\ref{eq:empirical-align-objective} is the empirical counterpart of the upper bound in Eq.~\ref{eq:sandwich-bound}, which in turn bounds the score-distribution discrepancy associated with $\Delta_{P,Q}$ in Eq.~\ref{eq:density-bound}. Therefore, minimizing Eq.~\ref{eq:empirical-align-objective} reduces an empirical upper bound on $\Delta_{P,Q}$. By Eq.~\ref{eq:cov_guarantee}, this tightens the lower bound on target coverage and helps recover the desired coverage level.

\section{Experiments}
\label{sec:exp}
\newcommand{\g}[1]{\cellcolor{green!10}#1}
\newcommand{\orange}[1]{\cellcolor{gray!21}#1}

\begin{table}[t]
\caption{
Comparison of \ours{} with conformal prediction baselines, averaged over nine
datasets across three modalities and three conformity scores.
}
\label{tab:main}
\centering

\resizebox{\columnwidth}{!}{
\setlength{\tabcolsep}{2pt}
\renewcommand{\arraystretch}{1.2}
\begin{tabular}{l|l|c|c|ccc|ccc}
\toprule
\multirow{2}{*}{\textbf{Score}} &
\multirow{2}{*}{\textbf{Method}} &
\multirow{2}{*}{\textbf{ACA}$\uparrow$} &
\multirow{2}{*}{\makecell{\textbf{Coverage}\\\textbf{Valid?}}} &
\multicolumn{3}{c|}{$\boldsymbol{\alpha = 0.10}$} &
\multicolumn{3}{c}{$\boldsymbol{\alpha = 0.05}$} \\
\cmidrule(lr){5-7}\cmidrule(lr){8-10}
& & & &
\textbf{Cov.} & \textbf{Size}$\downarrow$ & \textbf{CCV}$\downarrow$ &
\textbf{Cov.} & \textbf{Size}$\downarrow$ & \textbf{CCV}$\downarrow$ \\
\midrule

\multirow{7}{*}{\textbf{LAC}}
& SCP
& 50.1
& \cmark
& 0.894\std{0.004} & 4.00 & 8.92
& 0.949\std{0.004} & 4.80 & 5.09 \\

& \orange{Weighted CP}
& \orange{63.9}
& \orange{\xmark}
& \orange{0.862\std{0.003}} & \orange{7.89} & \orange{76.67}
& \orange{0.869\std{0.004}} & \orange{7.89} & \orange{80.97} \\

& \orange{Adapt+SCP}
& \orange{63.9}
& \orange{\xmark}
& \orange{0.885\std{0.004}} & \orange{2.86} & \orange{5.70}
& \orange{0.940\std{0.003}} & \orange{3.48} & \orange{3.58} \\

& SCAT
& 55.2
& \cmark
& 0.898\std{0.003} & 3.30 & \underline{7.47}
& 0.952\std{0.003} & \underline{4.03} & \textbf{4.03} \\

& \g{\textbf{\ours (A)}}
& \g{\underline{63.4}{\scriptsize\textcolor{green!50!black}{$_{+8.2}$}}}
& \g{\cmark}
& \g{0.896\std{0.004}}
& \g{\textbf{2.17}{\scriptsize\textcolor{green!50!black}{$_{-1.13}$}}}
& \g{7.97{\scriptsize\textcolor{green!50!black}{$_{+0.50}$}}}
& \g{0.949\std{0.003}}
& \g{\textbf{3.64}{\scriptsize\textcolor{green!50!black}{$_{-0.39}$}}}
& \g{4.61{\scriptsize\textcolor{green!50!black}{$_{+0.58}$}}} \\

& \g{\textbf{\ours (LR)}}
& \g{57.4{\scriptsize\textcolor{green!50!black}{$_{+2.2}$}}}
& \g{\cmark}
& \g{0.916\std{0.004}}
& \g{3.62{\scriptsize\textcolor{green!50!black}{$_{+0.32}$}}}
& \g{7.91{\scriptsize\textcolor{green!50!black}{$_{+0.44}$}}}
& \g{0.951\std{0.003}}
& \g{4.61{\scriptsize\textcolor{green!50!black}{$_{+0.58}$}}}
& \g{4.16{\scriptsize\textcolor{green!50!black}{$_{+0.13}$}}} \\

& \g{\textbf{\ours (LP)}}
& \g{\textbf{63.9}{\scriptsize\textcolor{green!50!black}{$_{+8.7}$}}}
& \g{\cmark}
& \g{0.898\std{0.004}}
& \g{\underline{2.99}{\scriptsize\textcolor{green!50!black}{$_{-0.31}$}}}
& \g{\textbf{7.46}{\scriptsize\textcolor{green!50!black}{$_{-0.01}$}}}
& \g{0.951\std{0.003}}
& \g{\textbf{3.64}{\scriptsize\textcolor{green!50!black}{$_{-0.39}$}}}
& \g{\underline{4.09}{\scriptsize\textcolor{green!50!black}{$_{+0.06}$}}} \\

\midrule

\multirow{7}{*}{\textbf{APS}}
& SCP
& 50.1
& \cmark
& 0.896\std{0.003} & 4.05 & 8.70
& 0.953\std{0.003} & 4.87 & 4.82 \\

& \orange{Weighted CP}
& \orange{63.9}
& \orange{\xmark}
& \orange{0.862\std{0.003}} & \orange{7.89} & \orange{76.66}
& \orange{0.872\std{0.003}} & \orange{7.89} & \orange{79.98} \\

& \orange{Adapt+SCP}
& \orange{63.9}
& \orange{\xmark}
& \orange{0.885\std{0.004}} & \orange{2.86} & \orange{5.70}
& \orange{0.940\std{0.003}} & \orange{3.48} & \orange{3.58} \\

& SCAT
& 55.2
& \cmark
& 0.900\std{0.004} & 3.35 & \underline{7.18}
& 0.954\std{0.003} & \underline{4.08} & \underline{3.97} \\

& \g{\textbf{\ours (A)}}
& \g{\underline{63.4}{\scriptsize\textcolor{green!50!black}{$_{+8.2}$}}}
& \g{\cmark}
& \g{0.899\std{0.003}}
& \g{\textbf{2.27}{\scriptsize\textcolor{green!50!black}{$_{-1.08}$}}}
& \g{8.73{\scriptsize\textcolor{green!50!black}{$_{+1.55}$}}}
& \g{0.948\std{0.002}}
& \g{4.92{\scriptsize\textcolor{green!50!black}{$_{+0.84}$}}}
& \g{5.83{\scriptsize\textcolor{green!50!black}{$_{+1.86}$}}} \\

& \g{\textbf{\ours (LR)}}
& \g{57.4{\scriptsize\textcolor{green!50!black}{$_{+2.2}$}}}
& \g{\cmark}
& \g{0.907\std{0.004}}
& \g{3.59{\scriptsize\textcolor{green!50!black}{$_{+0.24}$}}}
& \g{7.82{\scriptsize\textcolor{green!50!black}{$_{+0.64}$}}}
& \g{0.951\std{0.002}}
& \g{4.83{\scriptsize\textcolor{green!50!black}{$_{+0.75}$}}}
& \g{4.19{\scriptsize\textcolor{green!50!black}{$_{+0.22}$}}} \\

& \g{\textbf{\ours (LP)}}
& \g{\textbf{63.9}{\scriptsize\textcolor{green!50!black}{$_{+8.7}$}}}
& \g{\cmark}
& \g{0.900\std{0.004}}
& \g{\underline{3.01}{\scriptsize\textcolor{green!50!black}{$_{-0.34}$}}}
& \g{\textbf{5.77}{\scriptsize\textcolor{green!50!black}{$_{-1.41}$}}}
& \g{0.954\std{0.002}}
& \g{\textbf{3.77}{\scriptsize\textcolor{green!50!black}{$_{-0.31}$}}}
& \g{\textbf{3.55}{\scriptsize\textcolor{green!50!black}{$_{-0.42}$}}} \\

\midrule

\multirow{7}{*}{\textbf{RAPS}}
& SCP
& 50.1
& \cmark
& 0.895\std{0.003} & 4.15 & 8.75
& 0.953\std{0.003} & 5.01 & 4.90 \\

& \orange{Weighted CP}
& \orange{63.9}
& \orange{\xmark}
& \orange{0.866\std{0.003}} & \orange{7.89} & \orange{76.66}
& \orange{0.869\std{0.003}} & \orange{7.89} & \orange{79.99} \\

& \orange{Adapt+SCP}
& \orange{63.9}
& \orange{\xmark}
& \orange{0.885\std{0.003}} & \orange{2.85} & \orange{5.71}
& \orange{0.939\std{0.003}} & \orange{3.50} & \orange{3.58} \\

& SCAT
& 55.2
& \cmark
& 0.903\std{0.004} & 3.40 & \underline{7.50}
& 0.954\std{0.003} & 4.18 & 4.57 \\

& \g{\textbf{\ours (A)}}
& \g{\underline{63.4}{\scriptsize\textcolor{green!50!black}{$_{+8.2}$}}}
& \g{\cmark}
& \g{0.897\std{0.003}}
& \g{\textbf{2.27}{\scriptsize\textcolor{green!50!black}{$_{-1.13}$}}}
& \g{8.75{\scriptsize\textcolor{green!50!black}{$_{+1.25}$}}}
& \g{0.948\std{0.002}}
& \g{\textbf{2.91}{\scriptsize\textcolor{green!50!black}{$_{-1.27}$}}}
& \g{5.89{\scriptsize\textcolor{green!50!black}{$_{+1.32}$}}} \\

& \g{\textbf{\ours (LR)}}
& \g{57.4{\scriptsize\textcolor{green!50!black}{$_{+2.2}$}}}
& \g{\cmark}
& \g{0.906\std{0.003}}
& \g{3.61{\scriptsize\textcolor{green!50!black}{$_{+0.21}$}}}
& \g{8.12{\scriptsize\textcolor{green!50!black}{$_{+0.62}$}}}
& \g{0.960\std{0.002}}
& \g{4.99{\scriptsize\textcolor{green!50!black}{$_{+0.81}$}}}
& \g{\underline{4.26}{\scriptsize\textcolor{green!50!black}{$_{-0.31}$}}} \\

& \g{\textbf{\ours (LP)}}
& \g{\textbf{63.9}{\scriptsize\textcolor{green!50!black}{$_{+8.7}$}}}
& \g{\cmark}
& \g{0.900\std{0.004}}
& \g{\underline{3.00}{\scriptsize\textcolor{green!50!black}{$_{-0.40}$}}}
& \g{\textbf{5.79}{\scriptsize\textcolor{green!50!black}{$_{-1.71}$}}}
& \g{0.953\std{0.002}}
& \g{\underline{3.77}{\scriptsize\textcolor{green!50!black}{$_{-0.41}$}}}
& \g{\textbf{3.57}{\scriptsize\textcolor{green!50!black}{$_{-1.00}$}}} \\

\bottomrule
\end{tabular}
}
\end{table}

\begin{table}[!ht]
\caption{
Comparison of \ours{} with transductive solvers.
$(S\!\rightarrow\!U)$ denotes supervised adaptation followed by unsupervised
transductive optimization, while (SS) denotes semi-supervised optimization
using the labeled support set.
}
\label{tab:transductive_baselines}
\centering
\resizebox{\columnwidth}{!}{
\setlength{\tabcolsep}{2pt}
\renewcommand{\arraystretch}{1.2}
\begin{tabular}{l|l|c|c|ccc|cc}
\toprule
\multirow{2}{*}{\textbf{Score}} &
\multirow{2}{*}{\textbf{Method}} &
\multirow{2}{*}{\textbf{ACA}$\uparrow$} &
\multirow{2}{*}{\makecell{\textbf{Coverage} \\ \textbf{Valid?}}} &
\multicolumn{3}{c|}{\textbf{$\alpha=0.10$}} &
\multicolumn{2}{c}{\textbf{Complexity}} \\
\cmidrule(lr){5-7} \cmidrule(lr){8-9}
& & & &
\textbf{Cov.}
& \textbf{Size}$\downarrow$
& \textbf{CCV}$\downarrow$
& \textbf{T}$\downarrow$
& \textbf{GPU}$\downarrow$ \\
\midrule

\multirow{9}{*}{\textbf{LAC}}
& SCP
& 50.1
& \cmark
& 0.894\std{0.003}
& 4.00
& 8.92
& 0.00
& -- \\

& \orange{TIM \cite{boudiaf2020information}}
& \orange{53.5\imp{+3.4}}
& \orange{\xmark}
& \orange{0.888\std{0.004}}
& \orange{3.96\imp{-0.04}}
& \orange{8.08\imp{-0.84}}
& \orange{1.12}
& \orange{0.6} \\

& \orange{TIM $(S\rightarrow U)$ \cite{boudiaf2020information}}
& \orange{57.0\imp{+6.9}}
& \orange{\xmark}
& \orange{0.878\std{0.004}}
& \orange{3.54\imp{-0.46}}
& \orange{8.69\imp{-0.23}}
& \orange{1.24}
& \orange{0.6} \\

& \orange{TIM (SS) \cite{boudiaf2020information}}
& \orange{63.0\imp{+12.9}}
& \orange{\xmark}
& \orange{0.863\std{0.004}}
& \orange{2.13\imp{-1.87}}
& \orange{7.54\imp{-1.38}}
& \orange{1.11}
& \orange{0.6} \\

& \orange{TransCLIP \cite{zanella2024boosting}}
& \orange{54.8\imp{+4.7}}
& \orange{\xmark}
& \orange{0.726\std{0.003}}
& \orange{2.16\imp{-1.84}}
& \orange{22.31\wor{+13.39}}
& \orange{0.47}
& \orange{1.1} \\

& \orange{TransCLIP $(S\rightarrow U)$ \cite{zanella2024boosting}}
& \orange{61.3\imp{+11.2}}
& \orange{\xmark}
& \orange{0.866\std{0.003}}
& \orange{2.05\imp{-1.95}}
& \orange{8.06\imp{-0.86}}
& \orange{0.83}
& \orange{1.1} \\

& \orange{TransCLIP (SS) \cite{zanella2024boosting}}
& \orange{54.0\imp{+3.9}}
& \orange{\xmark}
& \orange{0.863\std{0.004}}
& \orange{3.04\imp{-0.96}}
& \orange{7.41\imp{-1.51}}
& \orange{0.45}
& \orange{1.1} \\

& SCAT \cite{silva2025trustworthy}
& \underline{55.2}\imp{+5.1}
& \cmark
& 0.898\std{0.003}
& \underline{3.30}\imp{-0.70}
& \underline{7.47}\imp{-1.45}
& 1.04
& 0.6 \\

& \g{\textbf{\ours}}
& \g{\textbf{63.9}\imp{+13.8}}
& \g{\cmark}
& \g{0.898\std{0.004}}
& \g{\textbf{2.99}\imp{-1.01}}
& \g{\textbf{7.46}\imp{-1.46}}
& \g{1.43}
& \g{1.0} \\

\midrule

\multirow{9}{*}{\textbf{APS}}
& SCP
& 50.1
& \cmark
& 0.896\std{0.003}
& 4.05
& 8.69
& 0.00
& -- \\

& \orange{TIM \cite{boudiaf2020information}}
& \orange{53.5\imp{+3.4}}
& \orange{\xmark}
& \orange{0.887\std{0.004}}
& \orange{3.16\imp{-0.89}}
& \orange{7.47\imp{-1.22}}
& \orange{1.18}
& \orange{0.6} \\

& \orange{TIM $(S\rightarrow U)$ \cite{boudiaf2020information}}
& \orange{57.0\imp{+6.9}}
& \orange{\xmark}
& \orange{0.880\std{0.004}}
& \orange{3.60\imp{-0.45}}
& \orange{8.00\imp{-0.69}}
& \orange{1.58}
& \orange{0.6} \\

& \orange{TIM (SS) \cite{boudiaf2020information}}
& \orange{63.0\imp{+12.9}}
& \orange{\xmark}
& \orange{0.863\std{0.004}}
& \orange{2.13\imp{-1.92}}
& \orange{7.54\imp{-1.15}}
& \orange{1.22}
& \orange{0.6} \\

& \orange{TransCLIP \cite{zanella2024boosting}}
& \orange{54.8\imp{+4.7}}
& \orange{\xmark}
& \orange{0.733\std{0.003}}
& \orange{2.52\imp{-1.53}}
& \orange{21.78\wor{+13.09}}
& \orange{0.40}
& \orange{1.1} \\

& \orange{TransCLIP $(S\rightarrow U)$ \cite{zanella2024boosting}}
& \orange{61.3\imp{+11.2}}
& \orange{\xmark}
& \orange{0.866\std{0.003}}
& \orange{2.05\imp{-2.00}}
& \orange{8.26\imp{-0.43}}
& \orange{0.65}
& \orange{1.1} \\

& \orange{TransCLIP (SS) \cite{zanella2024boosting}}
& \orange{54.0\imp{+3.9}}
& \orange{\xmark}
& \orange{0.863\std{0.003}}
& \orange{2.13\imp{-1.92}}
& \orange{7.54\imp{-1.15}}
& \orange{0.42}
& \orange{1.1} \\

& SCAT \cite{silva2025trustworthy}
& \underline{55.2}\imp{+5.1}
& \cmark
& 0.900\std{0.004}
& \underline{3.35}\imp{-0.70}
& \underline{7.18}\imp{-1.51}
& 1.15
& 0.6 \\

& \g{\textbf{\ours}}
& \g{\textbf{63.9}\imp{+13.8}}
& \g{\cmark}
& \g{0.900\std{0.004}}
& \g{\textbf{3.01}\imp{-1.04}}
& \g{\textbf{5.77}\imp{-2.92}}
& \g{1.54}
& \g{1.0} \\

\bottomrule
\end{tabular}
}
\end{table}

\subsection{Experiment Settings}

\noindent
\textbf{Dataset and Foundation Models.}
Following \cite{silva2025trustworthy}, we conduct experiments on nine datasets across three medical imaging modalities: histology (NCT-CRC \cite{kather2018100}, SICAPv2 \cite{silva2020going} and SkinCancer \cite{kriegsmann2022deep}), ophthalmology (MESSIDOR \cite{decenciere2014feedback}, MMAC \cite{qian2024competition}, FIVES \cite{jin2022fives}), and chest X-ray (CheXpert \cite{irvin2019chexpert,wang2022medclip}, NIH-LT \cite{wang2017chestx,holste2022long}, COVID \cite{chowdhury2020can,rahman2021exploring}). For each modality, we employ a modality-specific pretrained foundation model: CONCH \cite{lu2024visual} for histology, FLAIR \cite{silva2025foundation} for ophthalmology, and CONVIRT \cite{zhang2022contrastive} for chest X-ray to ensure strong modality-specific representations. All datasets are treated as multi-class classification problems, with an average of eight classes per dataset.

\begin{figure}[t]
  \centering
  \includegraphics[width=.8\linewidth]{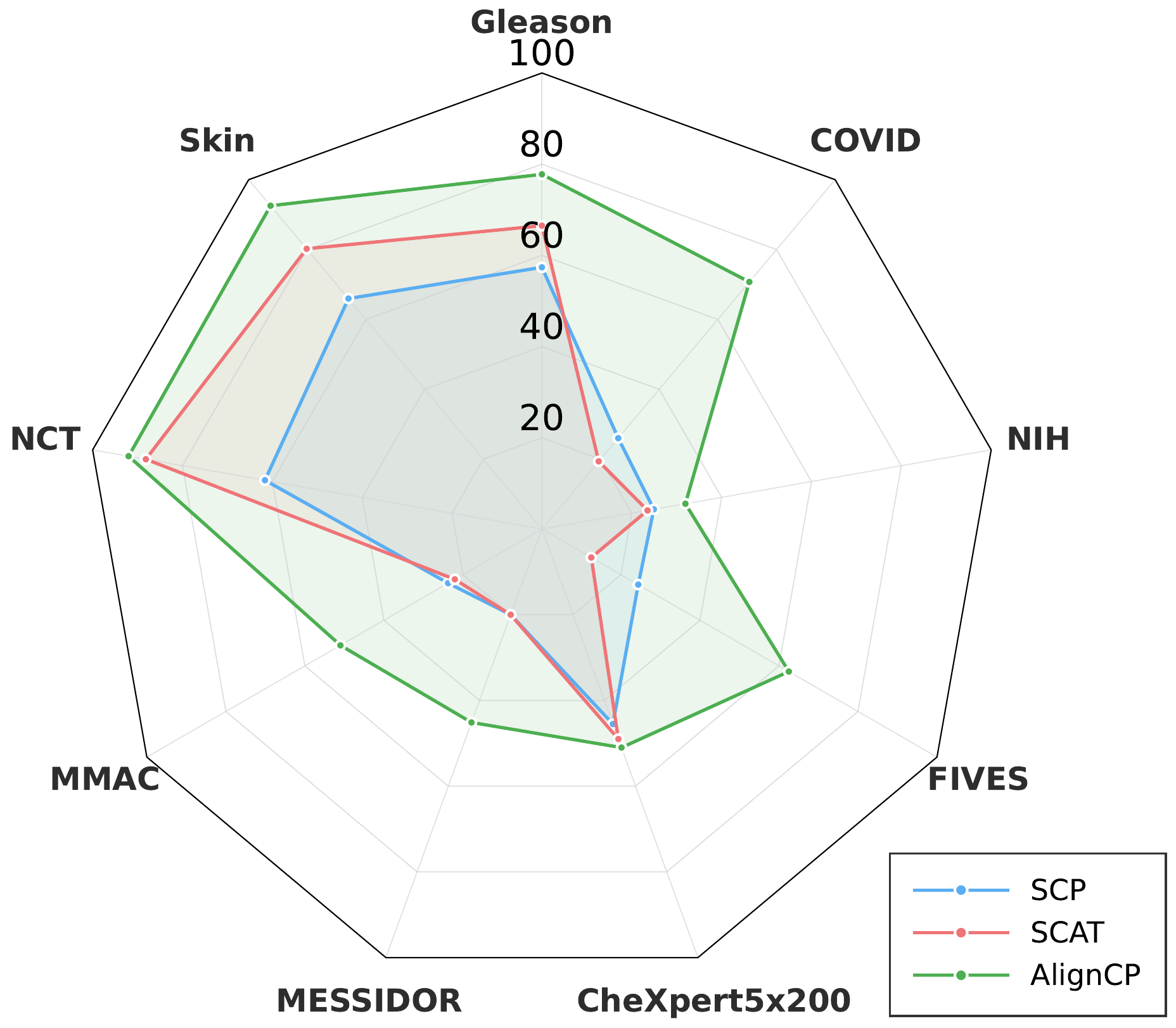}
  \caption{Dataset-wise ACA comparison across nine datasets for SCP, SCAT, and \ours.}
  \label{fig:aca}
\end{figure}

\noindent
\textbf{Implementation Details.}
For \ours, we first fine-tune the VLMs in a supervised manner on the calibration set using linear probing (LP), LoRA (LR) and residual adapter (A) strategy with linear probing be the main baseline, following \cite{gao2024clip,silva2024closer,hu2021lora}. To learn the importance weights for distribution alignment, we parameterize the weights as free learnable variables and optimize them using Adam with a learning rate of $10^{-5}$ for 300 epochs.
The weights are initialized uniformly and normalized via a softmax function to ensure they form a valid probability distribution (i.e., sum to one). The training objective corresponds to the loss defined in Equation \ref{eq:empirical-align-objective}.
All experiments are conducted on a single NVIDIA RTX 3090 GPU (24GB). Each configuration is repeated 100 times with different random seeds to ensure statistical robustness.

\noindent
\textbf{Conformal prediction.}
For conformal prediction, we employ three nonconformity scores for multi-class classification: RAPS \cite{angelopoulos2020uncertainty}, LAC \cite{sadinle2019least}, and APS \cite{romano2020classification}. For RAPS, we set the hyperparameters to k=1 and $\lambda = 0.001$. Experiments are conducted at two level $\alpha = 0.1$ and $\alpha = 0.05$

\noindent
\textbf{Metric.}
For evaluation metrics, following \cite{silva2025conformal}, we report balanced classwise accuracy (ACA), empirical coverage (“Cov.”), average prediction set size (“Size”), and class-conditioned coverage gap (“CCV”) \cite{ding2023class}.
To assess computational efficiency, we additionally report GPU memory consumption (“GPU”, in GB) and inference time (“T”, in seconds).
We consider a method undercovered when its empirical coverage falls more than 0.01 below the target coverage level. In all tables, gray rows indicate undercoverage. The Coverage Valid column marks whether a method satisfies this coverage criterion. Among coverage-valid methods, \textbf{bold} and \underline{underlined} values denote the best and second-best results, respectively.

\subsection{Experiment Result}

\noindent
\textbf{Overall Performance.}
Table~\ref{tab:main} compares \ours{} with SCP, Adapt+SCP, Weighted CP~\cite{barber2023conformal}, and SCAT~\cite{silva2025trustworthy} across nine datasets and three nonconformity scores. Adapt+SCP uses the same supervised adaptation as \ours{} but directly applies standard SCP afterward, without the proposed score-distribution alignment. This makes it the most direct baseline for isolating the effect of our calibration step.

\ours{} preserves the accuracy gains from supervised adaptation while maintaining coverage close to the target at both $\alpha=0.10$ and $\alpha=0.05$. In contrast, Adapt+SCP achieves the same ACA as the linear-probe variant of \ours{} (63.9) but consistently undercovers, while Weighted CP shows even more severe undercoverage. Among coverage-valid methods, \ours{} also yields smaller prediction sets than SCP and SCAT with competitive CCV. The consistent performance across Adapter, LoRA, and Linear Probe further shows that the proposed alignment is robust to the adaptation strategy. Overall, \ours{} provides a better balance of accuracy, coverage validity, set efficiency, and class-conditional reliability under target shift.

\noindent
\textbf{Comparison with Transductive Baselines.}
Table~\ref{tab:transductive_baselines} compares \ours{} with TIM, TransCLIP, and SCAT under LAC and APS. Although transductive adaptation improves classification accuracy, most TIM and TransCLIP variants suffer from undercoverage. In contrast, \ours{} achieves the best accuracy while maintaining target coverage. Among methods that preserve coverage, \ours{} also produces the most efficient prediction sets with the lowest CCV, while incurring computational cost comparable to existing transductive solvers. These results show that \ours{} provides a stronger trade-off between adaptation performance, uncertainty reliability, and prediction-set efficiency.



\noindent
\textbf{Effect of Calibration Shots.}
\begin{figure}[t]
  \centering
  \includegraphics[width=\linewidth]{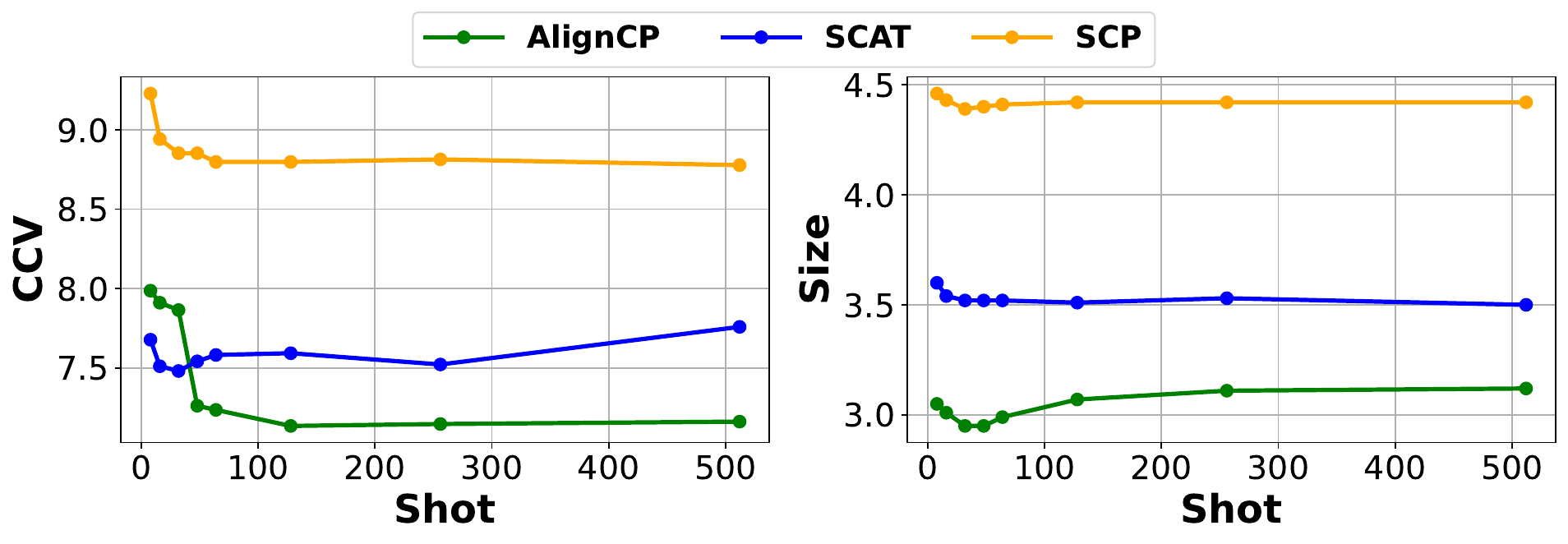}
  \caption{CCV and average prediction set size using LAC at $\alpha = 0.1$ for SCAT, SCP, and \ours across different labeled shots. Lower is better.}
  \label{fig:abl}
\end{figure}
Figure~\ref{fig:abl} studies the effect of calibration-set size on prediction-set size and class-conditioned coverage variation (CCV). Across all shot regimes, \ours{} consistently produces smaller prediction sets than SCP and SCAT, demonstrating improved set efficiency.

For CCV, \ours{} is slightly worse than SCAT in the extremely low-shot regime, where classwise estimates are noisy. As more calibration samples become available, however, \ours{} quickly achieves the lowest CCV and remains consistently better than both baselines. Overall, \ours{} benefits effectively from additional calibration data, yielding both compact prediction sets and improved classwise coverage balance.

\begin{figure}[t]
  \centering
  \includegraphics[width=\linewidth]{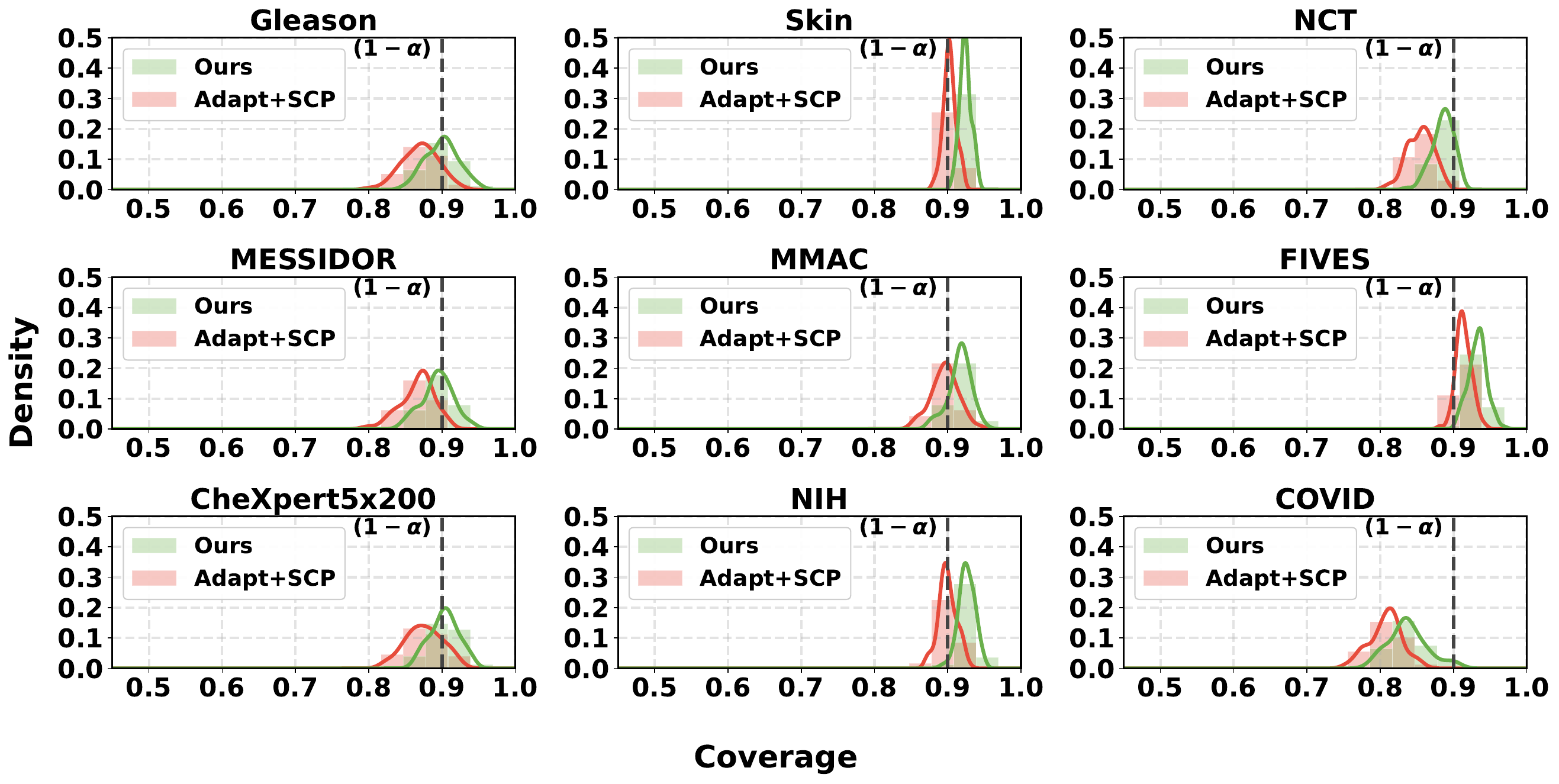}
  \caption{Coverage distributions across nine datasets at the target level $1-\alpha$. The dashed vertical line denotes the desired coverage. Compared with Adapt+SCP, which applies standard SCP after supervised adaptation without our calibration alignment, \ours{} produces coverage distributions closer to the target level across most datasets, reducing the undercoverage introduced by adaptation.}
  \label{fig:cov}
\end{figure}

\subsection{Coverage Distribution Analysis}
\label{sec:coverage_distribution}

Figure~\ref{fig:cov} isolates the effect of the proposed calibration alignment by comparing \ours{} with Adapt+SCP under the same supervised adaptation setting. Both methods use the same labeled support set and the same adapted predictor; therefore, they share the same predictive model and accuracy. The only difference is the conformal calibration step. Adapt+SCP applies standard SCP after adaptation, whereas \ours{} additionally learns the calibration weights by minimizing the empirical alignment objective in Eq.~\ref{eq:empirical-align-objective}.


Figure~\ref{fig:cov} provides direct empirical evidence for this effect. Adapt+SCP frequently produces coverage distributions shifted below the target level, with clear undercoverage on Gleason, NCT, MESSIDOR, CheXpert5x200, NIH, and COVID. In contrast, \ours{} shifts the distributions toward the target across the nine datasets. The distributions are centered close to the target on Gleason, MESSIDOR, MMAC, and CheXpert5x200, while Skin, FIVES, and NIH show slightly conservative coverage. Even on COVID, where the remaining shift is larger, \ours{} substantially reduces the undercoverage relative to Adapt+SCP.

Since the two methods use the same supervised predictor, these differences cannot be attributed to improved classification accuracy. They directly reflect the effect of minimizing Eq.~\ref{eq:empirical-align-objective}. Together, Fig.~\ref{fig:cov} and Table~\ref{tab:main} show that supervised adaptation alone improves prediction but can degrade conformal validity, while the proposed score-distribution alignment preserves the adapted predictor and brings empirical coverage substantially closer to the target level.

\begin{table}[t]
\centering
\caption{
Results on natural-image datasets. Entries report Coverage / Size / CCV.}

\label{tab:natural_image}
\resizebox{\linewidth}{!}{
\setlength{\tabcolsep}{2pt}
\renewcommand{\arraystretch}{1.2}
\scriptsize
\begin{tabular}{llc|cc|cc}
\toprule
\textbf{Dataset}
& \textbf{Method}
& \textbf{Acc.}
& \multicolumn{2}{c|}{\textbf{LAC}}
& \multicolumn{2}{c}{\textbf{APS}} \\
\cmidrule(lr){4-5}
\cmidrule(lr){6-7}
&
&
& $\alpha=.10$ & $\alpha=.05$
& $\alpha=.10$ & $\alpha=.05$ \\
\midrule

\multirow{4}{*}{ImageNet}
& SCP
& \underline{59.37}
& 0.895 / \textbf{6.67} / \underline{8.0}
& 0.946 / \textbf{17.58} / \textbf{5.0}
& 0.896 / \underline{19.78} / \textbf{6.0}
& 0.949 / \underline{42.29} / \textbf{3.0} \\

& Adapt+SCP
& \textbf{69.43}
& \cellcolor{gray!20}0.867 / 5.48 / 6.0
& \cellcolor{gray!20}0.932 / 14.92 / 5.0
& \cellcolor{gray!20}0.869 / 9.66 / 6.0
& \cellcolor{gray!20}0.931 / 20.52 / 5.0 \\

& SCA-T
& 58.86
& 0.890 / 26.23 / \underline{8.0}
& \cellcolor{gray!20}0.903 / 35.90 / 5.0
& 0.895 / 43.60 / \underline{8.0}
& \cellcolor{gray!20}0.922 / 140.78 / 5.0 \\

& \textbf{AlignCP}
& \textbf{69.43}
& 0.899 / \underline{8.48} / \textbf{6.0}
& 0.952 / \underline{22.72} / \textbf{5.0}
& 0.898 / \textbf{11.77} / \textbf{6.0}
& 0.952 / \textbf{28.84} / \underline{5.0} \\

\midrule

\multirow{4}{*}{CIFAR-10}
& SCP
& 88.32
& 0.892 / \underline{1.01} / \underline{6.5}
& 0.948 / \underline{1.30} / \underline{3.8}
& 0.917 / 1.60 / 2.2
& 0.959 / 1.96 / 1.8 \\

& Adapt+SCP
& \textbf{93.50}
& \cellcolor{gray!20}0.868 / 0.94 / 5.7
& \cellcolor{gray!20}0.912 / 1.06 / 5.3
& 0.910 / 1.35 / \textbf{1.4}
& 0.957 / 1.64 / \textbf{1.1} \\

& SCA-T
& \underline{89.17}
& \cellcolor{gray!20}0.880 / 0.93 / 4.7
& \cellcolor{gray!20}0.930 / 1.04 / 2.4
& 0.904 / \textbf{1.05} / 2.75
& 0.947 / \textbf{1.16} / 1.9 \\

& \textbf{AlignCP}
& \textbf{93.50}
& 0.891 / \textbf{0.97} / \textbf{5.3}
& 0.950 / \textbf{1.27} / \textbf{2.5}
& 0.906 / \underline{1.33} / \underline{1.6}
& 0.956 / \underline{1.63} / \underline{1.6} \\

\bottomrule
\end{tabular}
}
\end{table}
\begin{table}[t]
\centering
\caption{
Sensitivity of \ours{} to initialization, learning rate, and optimization iterations
under LAC at $\alpha=0.10$. Results for each hyperparameter are averaged over
the remaining settings.
}
\label{tab:weight_sensitivity}

\resizebox{\columnwidth}{!}{
\setlength{\tabcolsep}{14pt}
\renewcommand{\arraystretch}{1.08}
\begin{tabular}{llccc}
\toprule
\textbf{Hyperparameter}
& \textbf{Setting}
& \textbf{Cov.}
& \textbf{Size}$\downarrow$
& \textbf{CCV}$\downarrow$ \\
\midrule

\multirow{2}{*}{Initialization}
& Random
& 0.899 & 3.08 & 7.950 \\
& Uniform
& 0.898 & 3.01 & 7.849 \\

\midrule

\multirow{3}{*}{Learning Rate}
& $10^{-8}$
& 0.899 & 3.045 & 7.9 \\
& $10^{-7}$
& 0.898 & 3.045 & 7.9 \\
& $10^{-6}$
& 0.899 & 3.045 & 7.9 \\

\midrule

\multirow{3}{*}{Iterations}
& 100
& 0.898 & 3.045 & 7.9 \\
& 500
& 0.898 & 3.045 & 7.9 \\
& 1000
& 0.898 & 3.045 & 7.9 \\

\bottomrule
\end{tabular}
}
\end{table}








\begin{table}[t]
\centering
\caption{
Statistical comparison of ACA over 100 matched runs.
We report the mean paired difference, paired-bootstrap 95\% confidence
interval, and two-sided Wilcoxon signed-rank test.
}
\label{tab:statistical_test}
\resizebox{\columnwidth}{!}{
\setlength{\tabcolsep}{10pt}
\renewcommand{\arraystretch}{1.08}
\begin{tabular}{lccc}
\toprule
\textbf{Comparison}
& $\boldsymbol{\Delta}$\textbf{ACA}
& \textbf{95\% CI}
& $\boldsymbol{p}$\textbf{-value} \\
\midrule
\ours{} vs. SCP
& +13.815
& [13.807, 13.821]
& $3.9\times10^{-18}$ \\
\ours{} vs. SCAT
& +8.729
& [8.720, 8.735]
& $3.9\times10^{-18}$ \\
\bottomrule
\end{tabular}
}

\end{table}

\subsection{Natural-Image Generalization.}
Table~\ref{tab:natural_image} evaluates \ours{} with CLIP on ImageNet \cite{deng2009imagenet} and CIFAR-10 \cite{krizhevsky2009learning}. \ours{} preserves the accuracy gains from supervised adaptation while maintaining coverage close to the target levels, whereas Adapt+SCP frequently undercovers despite similar accuracy. Across both datasets and conformal scores, \ours{} also maintains competitive prediction-set size and CCV. These results confirm that the proposed score-distribution alignment is not specific to medical imaging and generalizes to standard natural-image benchmarks.

\subsection{Sensitivity Analysis.}
Table~\ref{tab:weight_sensitivity} evaluates sensitivity to weight initialization, learning rate, and optimization iterations. Coverage, set size, and CCV vary only marginally across all settings, including random versus uniform initialization. The results also remain essentially unchanged across learning rates and iteration counts, indicating that the learned weighting and resulting conformal predictions are stable to the optimization configuration.

\subsection{Statistical Significance.}
Table~\ref{tab:statistical_test} reports paired statistical comparisons over 100 matched runs. \ours{} improves ACA over SCP and SCAT by $13.815$ and $8.729$, respectively. The corresponding 95\% confidence intervals are narrow and remain well above zero, while the two-sided Wilcoxon signed-rank tests \cite{wilcoxon1992individual} yield $p=3.9\times10^{-18}$ for both comparisons. These results confirm that the observed accuracy gains are consistent and statistically significant across repeated runs.





\section{Conclusion}


In this work, we proposed \ours, a principled framework for supervised few-shot conformal transfer in medical VLMs. \ours addresses the score-level mismatch introduced by supervised adaptation by learning a reweighted calibration score distribution that better matches the target-domain behavior. This allows the model to benefit from few-shot fine-tuning while reducing the resulting coverage gap. Experiments across diverse medical tasks and modalities show that \ours improves adaptation performance, maintains reliable coverage, produces smaller prediction sets, and reduces class-conditioned coverage disparities. These results highlight score-distribution alignment as an effective way to combine supervised adaptation with conformal reliability.

\noindent
\textbf{Limitation.}
The coverage gap bound depends on surrogate score distributions and can loosen when they poorly approximate the test distribution. In extremely low-shot settings, limited calibration data may also make the bounds less stable and increase CCV. Future work will improve score modeling for more reliable guarantees in data-scarce regimes.

\newpage
{
    \small
    \bibliographystyle{ieeenat_fullname}
    \bibliography{main}
}

\end{document}